\documentclass{svproc}
\usepackage{graphicx}
\usepackage[dvipsnames]{xcolor}
\usepackage{colortbl}
\usepackage{amssymb}
\usepackage{empheq}
\usepackage{mathrsfs}
\usepackage{enumerate}
\usepackage{tikz}
\usepackage{upgreek}
\usepackage{tipa}
\usepackage{gensymb}
\usepackage{multicol}
\usepackage{textcomp}
\usepackage{tabularx}
\usepackage[]{algorithm2e}
\usepackage{minted}
\usepackage[toc,page]{appendix}
\usepackage{hyperref}
\usepackage{todonotes}
\usepackage{adjustbox}
\usepackage{longtable}
\usepackage{pdflscape} 
\usepackage{ulem}
\usepackage{cancel}
\usepackage{fancyhdr}
\usepackage{subcaption}
\usepackage{wrapfig}
\usepackage{libertinus}
\usepackage{soul}
\usepackage{multirow}
\usepackage{multicol}
\usepackage{booktabs}
\usepackage{lineno}
\usepackage[T1]{fontenc}
\usepackage{algpseudocode}
\usepackage{url}

\begin{document}
\mainmatter              
%
\title{TinyIceNet: Low-Power SAR Sea Ice Segmentation for On-Board FPGA Inference}
\titlerunning{TinyIceNet}  
%
\author{Mhd Rashed Al Koutayni \and Mohamed Selim \and Gerd Reis \and Alain Pagani \and Didier Stricker}
\authorrunning{Al Koutayni et al.} 
\institute{
German Research Center for Artificial Intelligence, DFKI, 
67663 Kaiserslautern, Germany\\
\{mhd\_rashed.al\_koutayni, mohamed.selim, gerd.reis, alain.pagani, didier.stricker\}@dfki.de
}

\maketitle              

\begin{abstract}
Accurate sea ice mapping is essential for safe maritime navigation in polar regions, where rapidly changing ice conditions require timely and reliable information. While Sentinel-1 Synthetic Aperture Radar (SAR) provides high-resolution, all-weather observations of sea ice, conventional ground-based processing is limited by downlink bandwidth, latency, and energy costs associated with transmitting large volumes of raw data. 
On-board processing, enabled by dedicated inference chips integrated directly within the satellite payload, offers a transformative alternative by generating actionable sea ice products in orbit. 
In this context, we present TinyIceNet, a compact semantic segmentation network co-designed for on-board Stage of Development (SOD) mapping from dual-polarized Sentinel-1 SAR imagery under strict hardware and power constraints. 
Trained on the AI4Arctic dataset, TinyIceNet combines SAR-aware architectural simplifications with low-precision quantization to balance accuracy and efficiency. 
The model is synthesized using High-Level Synthesis and deployed on a Xilinx Zynq UltraScale+ FPGA platform, demonstrating near-real-time inference with significantly reduced energy consumption. 
Experimental results show that TinyIceNet achieves 75.216\% F1 score on SOD segmentation while reducing energy consumption by 2$\times$ compared to full-precision GPU baselines, underscoring the potential of chip-level hardware–algorithm co-design for future spaceborne and edge AI systems.
\keywords{Sea Ice Segmentation, Synthetic Aperture Radar (SAR), Quantized Neural Network, FPGA SoC, Low-Power Edge AI, On-Board Processing.}
\end{abstract}
\section{Introduction}
\label{section:introduction}
Sea ice monitoring is essential for maritime safety in polar regions, where rapidly changing ice conditions can pose severe risks to navigation.
Among the key parameters, the stage of development (SOD) is especially important, as it reflects ice thickness and maturity.
Accurate SOD information enables vessels to avoid hazardous areas, supports tactical decision-making during operations, and informs long-term planning such as voyage routing and ice forecasting.

Traditionally, SOD maps were produced through manual interpretation of satellite imagery.
While effective, this approach is slow, labor-intensive, and dependent on expert analysts, leading to delays in chart updates that can leave ships exposed to sudden changes in ice conditions.

Advancements in Synthetic Aperture Radar (SAR), particularly from the Sentinel-1 mission, have greatly improved sea ice observation under all-weather and day--night conditions.
However, converting SAR imagery into reliable SOD products remains computationally demanding.
While deep learning methods have demonstrated strong performance for sea ice segmentation, most existing approaches prioritize accuracy using large models and multi-modal inputs executed on ground-based computing infrastructures, limiting their suitability for time-critical and resource-constrained applications.

A promising direction is on-board satellite processing, where SAR data is analyzed directly in orbit.
Instead of downlinking vast volumes of raw imagery, satellites can generate SOD products in near-real-time, reducing bandwidth requirements and latency while delivering actionable information much faster.
Despite its clear operational advantages, on-board processing introduces strict constraints on power consumption, memory footprint, and numerical precision, which are not explicitly addressed by most existing sea ice segmentation pipelines.

In this work, we introduce \textit{TinyIceNet}, a lightweight convolutional neural network (CNN) specifically designed for SOD classification and segmentation from Sentinel-1 SAR imagery.
Rather than treating model compression and hardware deployment as post-processing steps, TinyIceNet follows a hardware-aware co-design approach in which network architecture, numerical precision, and deployment constraints are jointly considered.
Built on a simplified U-Net architecture and trained on the AI4Arctic dataset, TinyIceNet is optimized for low energy consumption and efficient deployment without compromising predictive accuracy.

A key focus of this work is the deployment of TinyIceNet on field-programmable gate arrays (FPGAs), which offer several advantages for spaceborne applications.
FPGAs provide high parallelism, reconfigurability, and low power consumption, making them well suited for processing large volumes of SAR data under tight energy and memory constraints.
Using High-Level Synthesis (HLS), we implement TinyIceNet on the Xilinx ZCU102 platform, demonstrating the feasibility of near-real-time SOD inference in resource-limited environments such as satellites and on-board ship systems.

Experimental results show that TinyIceNet achieves up to 75.216\% F1 accuracy on SOD segmentation, while reducing energy consumption by up to $2\times$ compared to full-precision GPU baselines.
The main contributions of this paper are as follows:
\begin{itemize}
    \item A compact SAR-specific segmentation architecture (TinyIceNet) that demonstrates how aggressive architectural simplification of U-Net–style models can preserve segmentation accuracy for low-texture, speckle-dominated SAR imagery, providing a transferable design principle for remote sensing models under tight resource constraints.
    
    \item A systematic quantization study showing that post-training quantization is insufficient for low-bit-width SAR inference, while quantization-aware training enables stable 8-bit deployment with negligible accuracy loss, offering practical guidance for precision selection in spaceborne and edge AI systems.
    
    \item An end-to-end hardware–algorithm co-design workflow that maps a SAR segmentation model to FPGA using HLS and DeepEdgeSoC, illustrating how co-optimizing network structure and numerical precision enables energy-efficient on-board inference beyond platform-specific performance tuning.
\end{itemize}

The rest of this paper is structured as follows: Section~\ref{section:related_works} reviews related work on sea ice monitoring and efficient deep learning models.
Section~\ref{section:data} introduces the AI4Arctic dataset used in this work.
Section~\ref{section:proposed_approach} describes the architecture of TinyIceNet, the training methodology, and the edge deployment process.
Section~\ref{section:experiments_results} presents experimental results and analysis.
Finally, Section~\ref{section:conclusion_future_works} concludes the paper and discusses directions for future research.
 
\section{Related Work}
\label{section:related_works}
SOD segmentation from SAR imagery has been explored through a range of approaches that can broadly be grouped into classical machine learning methods, deep learning–based segmentation models, and multi-task or large-scale benchmarking frameworks. Most studies rely on dual-polarized Sentinel-1 SAR data, although they differ substantially in target classes, spatial resolution, seasonal coverage, and evaluation protocols, which complicates direct performance comparison.

Early work predominantly employed feature-engineered pipelines, combining handcrafted texture descriptors with conventional classifiers. Park et al.~\cite{park2020classification} extracted Haralick texture features from HH and HV channels and used a Random Forest classifier to distinguish three coarse SOD-related classes in winter conditions, achieving approximately 87\% accuracy. While effective for ice charting at a coarse level, such methods require extensive feature computation and post-processing, with reported runtimes ranging from tens of minutes to hours per scene. These characteristics limit their scalability and suitability for fine-grained or near-real-time SOD mapping.

More recent studies adopt end-to-end deep learning models, typically based on CNN encoder–decoder architectures, to improve segmentation accuracy and robustness. Ren et al.~\cite{ren2021development} proposed a ResNet-based U-Net with attention mechanisms for binary ice–water segmentation, reporting improved intersection-over-union scores over a baseline U-Net. Kruk et al.~\cite{kruk2020proof} used DenseNet architectures to classify open water, first-year ice, and multi-year ice, achieving 94.0\% overall accuracy. Huang et al.~\cite{huang2024deep} introduced a dual-branch U-Net that fuses SAR intensity and texture features for multi-class ice segmentation. While these approaches demonstrate strong accuracy, they typically target offline analysis and rely on deep backbones with high computational and memory demands, making them unsuitable for embedded or on-board deployment.

A complementary line of work focuses on multi-task learning, where SOD is predicted jointly with other sea ice parameters. The MMSeaIce framework~\cite{chen2023mmseaice,chen2024mmseaice} employs a multi-task U-Net to estimate SOD alongside sea ice concentration (SIC) and floe size (FLOE) using SAR, passive microwave, and auxiliary geophysical inputs. Their ablation studies show that restricting inputs to SAR alone reduces SOD performance, yet still yields competitive F1 scores, highlighting SAR as a viable standalone modality. However, inference is evaluated exclusively on GPUs, with full-scene runtimes on the order of seconds, and hardware efficiency or low-power deployment is not addressed.

Large-scale initiatives such as AutoICE~\cite{stokholm2024autoice} and AI4SeaIce~\cite{stokholm2022ai4seaice,stokholm2023ai4seaice} emphasize benchmark-driven evaluation, input selection, and task separation strategies to improve generalization across regions and ice regimes. Additional studies investigate hierarchical classification pipelines and optimal input variable selection for SAR-based ice mapping~\cite{chen2023sea,chen2023influence,caceres2022landsat}. While these works provide valuable insights into model design and data usage, their focus remains on accuracy-oriented evaluation rather than on constraints imposed by on-board or edge inference.

Overall, existing literature demonstrates that deep learning significantly improves SAR-based SOD segmentation accuracy; however, most approaches prioritize predictive performance under offline or GPU-based settings. The question of how architectural simplification, numerical precision, and hardware constraints jointly affect SOD segmentation accuracy and feasibility for on-board deployment remains largely unexplored, motivating the approach presented in this work.
\section{Dataset and Preprocessing}
\label{section:data}
For this study, we use the AI4Arctic~\cite{ai4arctic_dataset} dataset to develop an automated sea ice mapping solution aimed at on-board satellite processing.
This dataset, used originally for the AutoICE challenge~\cite{stokholm2024autoice}, provides a rich collection of remotely sensed data for sea ice monitoring, including dual-polarized Sentinel-1 SAR imagery, AMSR2 microwave radiometer measurements, ERA5 weather reanalysis data, and expert-labeled sea ice charts following the World Meteorological Organization (WMO) classification system.
AI4Arctic was selected due to its high-quality expert annotations and its suitability for benchmarking operational sea ice mapping methods.
However, since our focus is on in-satellite inference, we restrict our input to the dual-polarized SAR data from Sentinel-1.
Sentinel-1 is equipped with a C-band SAR sensor (5.410~GHz, HH and HV polarizations) that captures high-resolution imagery in different weather and lighting conditions. 

The AI4Arctic dataset consists of 533 netCDF files, including 513 training files and 20 test files. 
Two versions of the dataset are available: a raw version and a ready-to-train version. 
In this work, we use the ready-to-train version to skip initial preparation steps and focus directly on model development.
Specifically, preprocessing includes downsampling The raw SAR data from Sentinel-1 to an 80-meter pixel resolution (roughly $5000\times5000$ pixels).
The pixel values are normalized to fall within a range of $[-1, 1]$, and special codes are used to handle missing data: NaN values in the SAR images are replaced with a value of 0, while areas outside the scope of the analysis (non-data or masked pixels) are assigned a value of 255.
Each scene is associated with a SOD chart map, which indicates the type and thickness of sea ice.
SOD is classified into five categories: 0 for open water, 1 for new ice, 2 for young ice, 3 for thin first-year ice, 4 for thick first-year ice, and 5 for old ice (older than one year).
\section{Proposed Approach}
\label{section:proposed_approach}
In this section, we present our methodology for near-real-time sea-ice SOD segmentation from dual-polarized SAR imagery.  
TinyIceNet, a compact encoder–decoder network, is trained on the AI4Arctic dataset and optimized for low-power, on-board inference.  
We detail its network design, training protocol, loss function and evaluation metric, quantization strategies, and FPGA deployment in the following subsections.

\subsection{Network Design}

\begin{figure}
    \centering
    \includegraphics[width=1\linewidth]{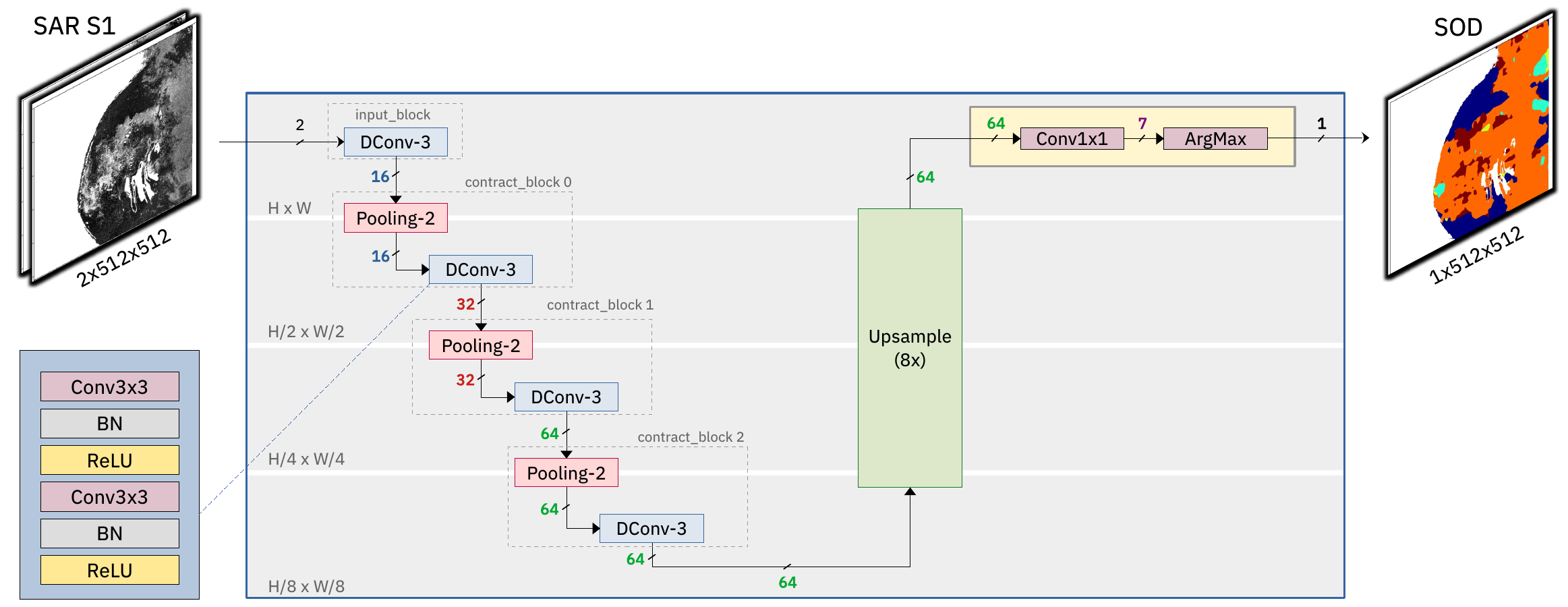}
    \caption{TinyIceNet Network Architecture, consisting of an encoder followed by upsample operation and a tiny segmentation head (Conv $1\times1$ and ArgMax). DConv-3 stands for Double Convolutional layer with a kernel size of $3$, as shown in the bottom left side.}
    \label{fig:Model}
\end{figure}

TinyIceNet (Figure~\ref{fig:Model}) predicts per-pixel SOD directly from HH and HV SAR inputs.
According to MMSeaIce~\cite{chen2023mmseaice,chen2023influence,chen2024comparative}, relying solely on SAR inputs for SOD prediction resulted in a notable accuracy drop by approximately 12.1\% compared to the full-feature model.
This degradation in performance can be attributed to the absence of brightness temperature data, which plays a vital role in enhancing model accuracy.
However, since the goal of this work is to develop a DNN for SAR on-board processing, we are restricted to using only SAR data and cannot incorporate additional satellite inputs.
TinyIceNet follows a compact encoder–decoder architecture designed for SOD segmentation from dual-polarized SAR imagery.
The encoder consists of three successive contracting blocks, where each block applies two standard $3\times3$ convolution layers with batch normalization and ReLU activation (denoted as DConv-3), followed by a pooling operation to progressively reduce spatial dimensions.
The convolution filters are configured as follows: the first layer has 16 filters, the second layer has 32 filters, and the next two layers has 64 filters.
The decoder avoids a complex expansive path by employing a single upsampling layer (×8) that restores the feature maps to the original spatial resolution.
A final $1\times1$ pointwise convolution projects the 64 feature channels into 7 output classes corresponding to different SOD categories.
An Argmax function is used to convert these raw SOD logits into class integers to form a final $1\times512\times512$ SOD map.
This corresponds to approx. 800-meter pixel resolution.
In fact, no skip connections between the encoder and decoder are used in our architecture.
This design choice follows the observations of Bahl \textit{et al.}~\cite{bahl2019low}, who demonstrated that while requiring substantial additional memory, skip connections may provide little to no benefit when the target scenes lack high-frequency spatial detail.
Analogously, SAR sea-ice SOD scenes exhibit limited high-frequency structure, making the removal of skip connections a natural choice that reduces memory and on-chip buffering requirements without degrading performance.
The computational complexity of TinyIceNet is 2.97~GMac for input size $2\times512\times512$ while having only 146.6~k parameters.

\subsection{Training Protocol}
The training process for TinyIceNet was conducted over a total of $200$~epochs, with each epoch consisting of $500$~iterations (steps).
All training experiments were carried out on a workstation equipped with an NVIDIA GeForce RTX 4090 GPU (24 GB memory), using PyTorch 2.4 with CUDA 12.5 support.
AI4Arctic dataset scenes were used and each input sample was resized to a dimension of 512$\times$512~pixels (800-meter pixel resolution) and saved to the disk for faster training.
TinyIceNet was designed to accept two input channels corresponding to the dual-polarized SAR data.
Stochastic Gradient Descent (SGD) optimizer is employed with a learning rate of 0.001, which was reduced gradually throughout training using a cosine annealing scheduler.
The input is processed in batches of 32 scenes.
The weight decay factor was set to 0.01 to prevent overfitting by regularizing the model's weights, and a momentum of 0.9 was used to accelerate convergence.

To enhance the model's robustness and improve generalization, data augmentation techniques were applied during training.
These included random horizontal and vertical flips, random rotations, and random scaling of the input images.
To guide the training process, 10 scenes from the training set were reserved as a validation set.
After each epoch, the model was evaluated on this validation subset, and a checkpoint was saved whenever the validation score surpassed that of the previously best-performing model.

\subsection{Loss Function and Evaluation Metric}
During the training of TinyIceNet, CrossEntropyLoss is used to optimize pixel-wise classification of SOD categories.
This loss compares predicted class probabilities with ground truth labels, guiding the model to improve accuracy by penalizing incorrect predictions.
The Cross-Entropy Loss for semantic segmentation is defined in Eq. \ref{eq:loss}.
\begin{equation}
\label{eq:loss}
\mathcal{L}_{CE} = - \frac{1}{N} \sum_{i=1}^{N} \sum_{c=1}^{C} y_i^c \log(\hat{p}_i^c)   
\end{equation}
where \( N \) is the total number of pixels, \( C \) is the number of classes, \( y_i^c \) is the ground truth indicator for pixel \(i\) and class \(c\), where \( y_i^c = 1 \) if pixel \(i\) belongs to class \(c\), and \( y_i^c = 0 \) otherwise.
\( \hat{p}_i^c \) is the predicted probability that pixel \(i\) belongs to class \(c\).
\( \log(\cdot) \) represents the natural logarithm.
Masked or ambiguous pixels have the default value of 255 in the dataset and therefore are excluded from the loss computation by setting the \texttt{ignore\_index} argument to 255 to avoid skewing results.

To evaluate the model's effectiveness, the F1 score is used.
The F1 score provides a balance between precision and recall, making it a suitable metric for evaluating classification tasks.
It also accounts for imbalances in the number of pixels across different classes. The F1 score is defined in Eq. \ref{eq:F1score}.
\begin{equation}
\label{eq:F1score}
F1 = 2 \frac{P \times R}{P + R}
\end{equation}
where $P = (TP)/(TP + FP)$ , and $R=TP/(TP + FN)$.
Here, $P$ is precision, $R$ is recall, $TP$ is the number of true positives, $FP$ is the number of false positives, and $FN$ is the number of false negatives.
Similarly to the loss function, any pixels that do not correspond to sea ice are excluded from the F1 score calculations.

\subsection{Network Optimization}
To enable efficient on-board inference on resource-limited satellite hardware, TinyIceNet is optimized through model quantization, a key technique for reducing memory footprint and computational load.
Quantization converts floating-point operations into low-precision fixed-point arithmetic, significantly improving speed and energy efficiency during deployment.
There are two principal quantization strategies: Post-Training Quantization (PTQ), which applies low-precision conversion after training, and Quantization-Aware Training (QAT), which simulates quantization effects during training to preserve accuracy.
In this work, we follow a pragmatic workflow: we initially apply PTQ since it is fast, requires no retraining, and provides an immediate assessment of the model’s resilience to reduced precision.
However, if aggressive quantization leads to substantial degradation, we transition to QAT, enabling the model to adapt to low-bit representations during training.
\subsection{FPGA Deployment Using High-Level Synthesis}
TinyIceNet is deployed on the AMD Xilinx ZCU102 evaluation board, which integrates the ZCU9EG SoC combining a quad-core ARM Cortex-A53 processor with FPGA fabric, using High-Level Synthesis (HLS) and the DeepEdgeSoC framework~\cite{ALKOUTAYNI2023100665}.
The ZCU9EG integrates a quad-core ARM Cortex-A53 processor with an FPGA fabric.
DeepEdgeSoC provides a library of modular C++ building blocks (e.g., convolutions, pooling, activation functions, and feature map buffers) that can be specialized at compile time using C++ template parameters.
Specifically, the number of processing elements, parallelization factor, data reuse, and quantization bitwidth can be easily customized without modifying the core algorithmic code.

\begin{figure}
    \centering
    \includegraphics[width=1\linewidth]{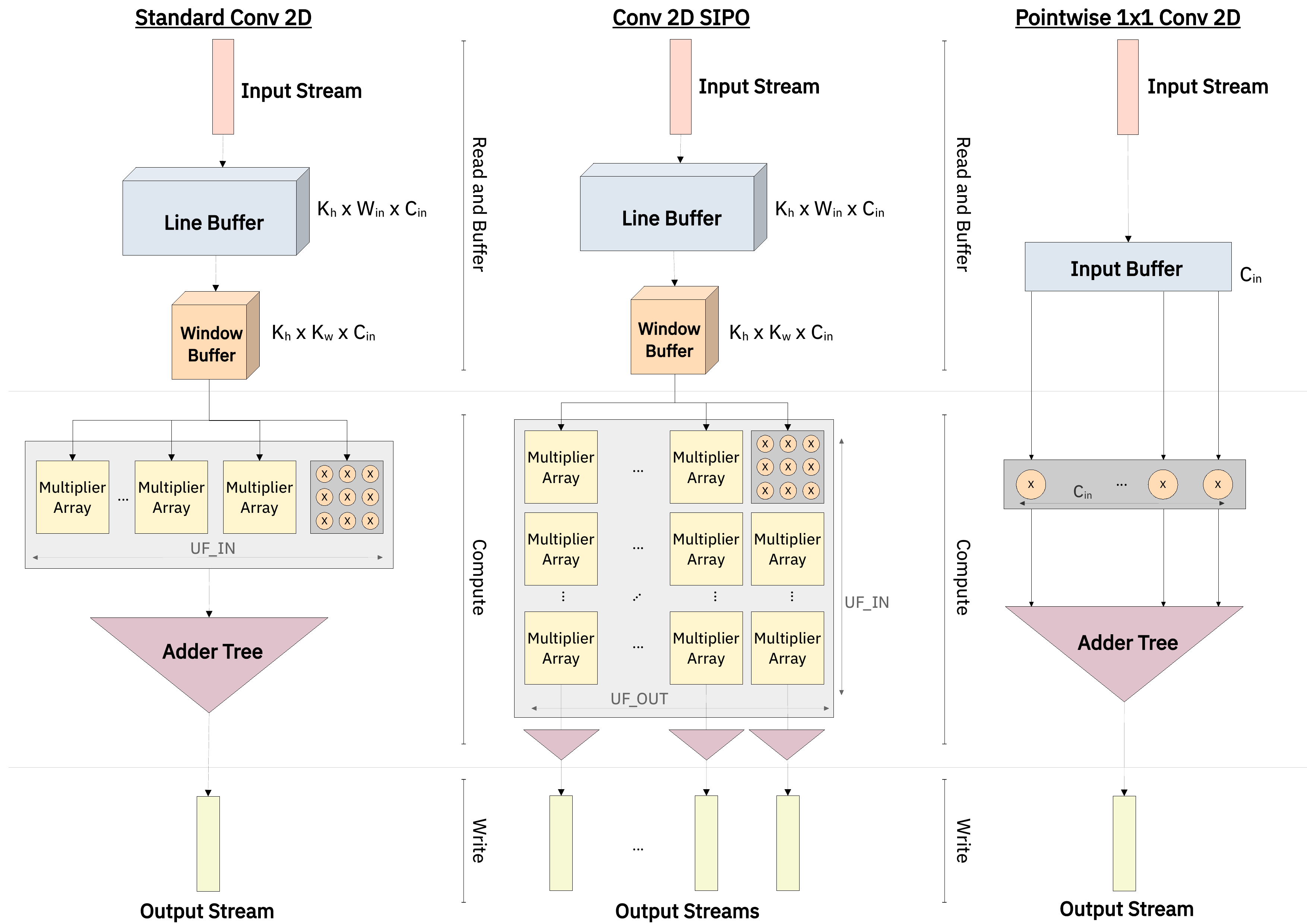}
    \caption{Overview of the streaming convolution architecture used in TinyIceNet.
        Each layer follows a unified pipeline consisting of (1) line-buffered input streaming, 
        (2) parallel MAC compute arrays scaled by input/output unrolling factors, 
        and (3) output streaming.
        Three variants are supported: standard Conv2D, Conv2D with Serial-In–Parallel-Out (SIPO) multi-output parallelism, 
        and pointwise $1 \times 1$ convolution.}
    \label{fig:conv_architecture}
\end{figure}

TinyIceNet uses a fully streaming convolution architecture optimized for FPGA implementation.
As can be seen in Figure~\ref{fig:Model}, TinyIceNet has 9 convolution layers, where only the last layer is a $1\times1$ pointwise convolution.
Each convolution layer follows a three-stage pipeline: \emph{Read and Buffer}, \emph{Compute}, and \emph{Write}.
As illustrated in Fig.~\ref{fig:conv_architecture}, incoming pixels are first cached in per-channel line buffers, from which a sliding $K_h \times K_w$ window is maintained. This enables continuous spatial reuse without off-chip memory access.
The compute module consists of a grid of MAC units whose replication is controlled by the input and output unrolling factors ($\mathrm{UF_{IN}}$, $\mathrm{UF_{OUT}}$).
In the standard Conv2D case, one output channel is produced per cycle.
However, when the standard Conv2D is a runtime bottleneck, we switch into the Serial-In–Parallel-Out (SIPO) variant which expands the MAC grid to emit multiple output channels in parallel from the same input window.
For the pointwise convolution, only channel buffering is required, and the operation reduces to a parallel dot-product followed by an adder tree.

\section{Experiments and Results}
\label{section:experiments_results}
This section evaluates TinyIceNet across accuracy, quantization robustness, runtime performance, and energy efficiency.
We compare full-precision, PTQ, and QAT variants using the AI4Arctic test split and assess their behavior on three hardware platforms: an RTX~4090 GPU, a Jetson AGX Xavier, and a ZCU102 FPGA SoC.
Quantitative metrics are complemented by qualitative visualizations to illustrate segmentation quality under different operating conditions.
\subsection{Experimental Setup}
All experiments are conducted using the AI4Arctic Sea Ice Dataset, focusing on pixel-wise SOD segmentation from dual-polarized Sentinel-1 SAR imagery.
The test split contains 20 scenes, and all models are evaluated using identical input preprocessing: normalization to the range $[-1,1]$, NaN replacement, and masking of invalid pixels (value 255).
Three model variants are compared: (i) the full-precision FP32 TinyIceNet, (ii) PTQ models with weight bit-widths ranging from 7--32 bits, and (iii) an 8-bit QAT model designed to recover accuracy under aggressive quantization. 
All models share the same architecture to ensure fair comparison.
Runtime and energy measurements are performed on three hardware platforms: an NVIDIA RTX~4090 workstation GPU, a Jetson AGX Xavier embedded GPU, and an AMD Xilinx ZCU102 FPGA SoC synthesized via HLS and DeepEdgeSoC.
For each platform, we measure: (i) inference throughput (fps), (ii) average power consumption, and (iii) energy per processed scene. 
FPGA resource utilization is obtained from post-implementation Vivado reports, including LUT, FF, BRAM, and DSP consumption.
To ensure consistency across full-precision and quantized model configurations, all reported results are based on end-to-end inference of full-resolution $512 \times 512$ scenes.

\subsection{Accuracy and Quantization Effects}

Table~\ref{tab:quant_f1} summarizes the impact of quantization on TinyIceNet's segmentation accuracy. 
The full-precision FP32 model achieves an F1 score of 75.168\%. PTQ shows a clear dependency on weight bitwidth: extremely low-precision weights (7--9 bits) severely degrade performance (13.19\%--43.52\%), whereas accuracy improves rapidly beyond 10 bits. Starting from 12 bits onward, PTQ stabilizes around 70--71\% F1, maintaining within 4--5\% of the baseline. The best PTQ performance is observed at 15 bits with 71.672\%.

\begin{figure}[t]
    \centering
    \includegraphics[width=1\linewidth]{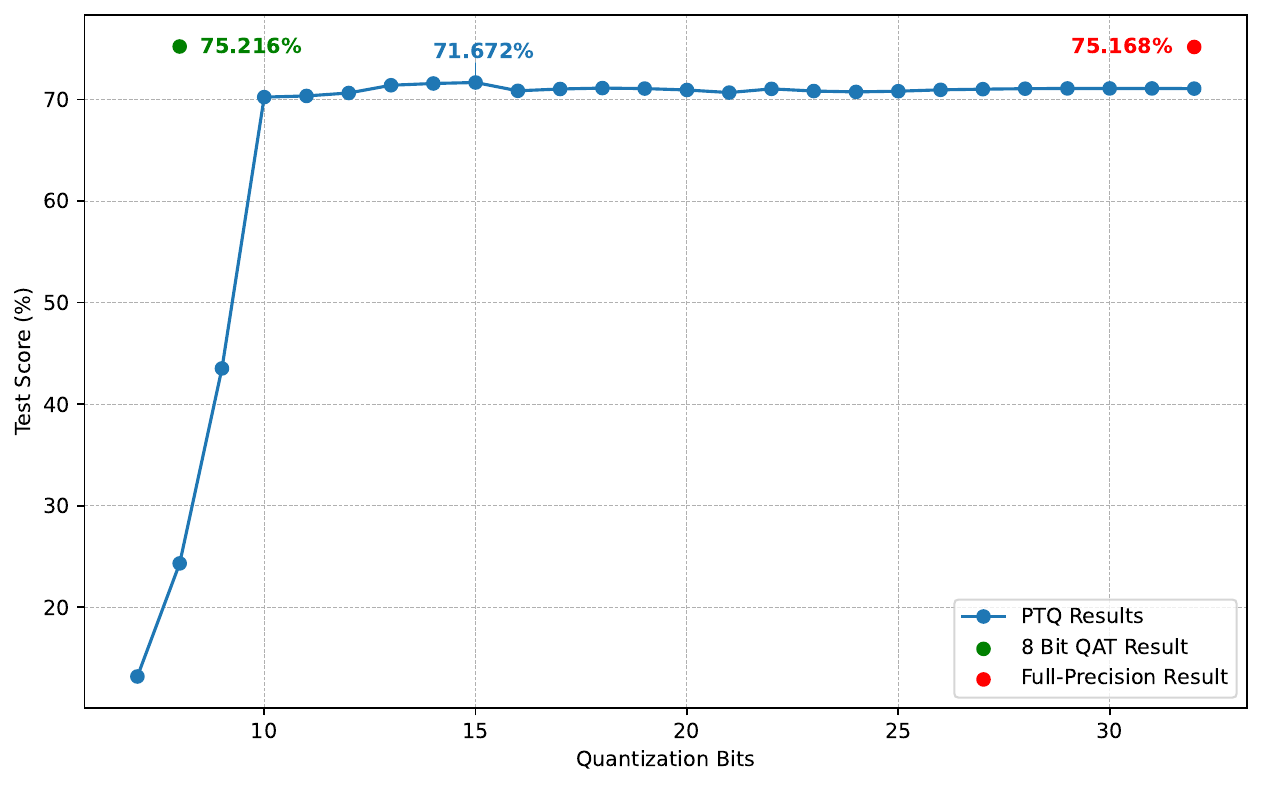}
    \caption{Accuracy curve: F1 score as a function of weight bitwidth for Full-Precision, QAT, and PTQ (7--32 bits). The line is drawn to illustrate the accuracy trend although the quantization steps and the corresponding F1 scores are discrete.}
    \label{fig:ptq_curve}
\end{figure}

In contrast, QAT at 8-bit weights fully recovers accuracy, achieving an F1 score of 75.216\%, matching and slightly surpassing the FP32 baseline.
The potential reason behind this slight improvement is the regularization effect that the quantization adds to the training during QAT.
Figure~\ref{fig:ptq_curve} illustrates the smooth trend of PTQ accuracy as a function of bitwidth.
It is worth mentioning that the test accuracy is calculated for resized scenes ($512\times512$).

\begin{table}
\centering
\caption{F1 score of TinyIceNet under different quantization schemes. The full-precision (FP32) implementation provides a comparison baseline. Different quantization bitwidths are experimented using Post-Training Quantization (PTQ). Quantization-Aware Training (QAT) at 8-bit weights fully recovers accuracy, matching and slightly surpassing the FP32 baseline.}
\label{tab:quant_f1}
\adjustbox{max width=\textwidth}{
\begin{tabular}{m{.5\textwidth} m{.5\textwidth}}
\toprule
\textbf{Configuration} & \textbf{F1 Score (\%)} \\
\midrule
FP32 (baseline) & 75.168 \\
PTQ (20-bit weights) & 70.934 \\
PTQ (15-bit weights) & 71.672 \\
PTQ (12-bit weights) & 70.639 \\
PTQ (10-bit weights) & 70.232 \\
PTQ (9-bit weights) & 43.519 \\
PTQ (8-bit weights) & 24.322 \\
PTQ (7-bit weights) & 13.189 \\
\textbf{QAT (8-bit weights)} & \textbf{75.216} \\
\bottomrule
\end{tabular}
}
\end{table}

\subsection{Runtime, Energy Efficiency and FPGA Resource Utilization}

To evaluate the practical suitability of TinyIceNet for edge and on-board inference, we benchmark its performance on three different hardware platforms: an NVIDIA RTX~4090 GPU, an NVIDIA Jetson AGX Xavier, and the AMD Xilinx ZCU102 SoC FPGA.
Table~\ref{tab:runtime} summarizes the achieved frame rates and the corresponding energy consumption per processed $512\times512$ scene.

The RTX~4090 provides the highest throughput at 764.8~fps, benefiting from its large number of CUDA cores and high memory bandwidth.
However, this performance comes at a relatively high energy cost of 228.7~mJ per scene, making it unsuitable for power-constrained environments such as satellites.

The Jetson AGX Xavier, designed for embedded applications, achieves 47.9~fps but exhibits the highest energy consumption (1218.5~mJ).
This result highlights the challenge of deploying deep learning models on embedded GPUs when both power and thermal budgets are limited.

In contrast, the ZCU102 FPGA achieves a lower throughput of 7~fps, yet offers a highly competitive energy profile, consuming only 113.6~mJ per scene.
Despite the lower frame rate, this energy efficiency makes the FPGA implementation compelling for on-board satellite processing, where power availability is severely restricted. 
Importantly, the achieved throughput remains sufficient for near–real-time processing of typical SAR acquisition rates.
\begin{table}
\centering
\caption{Inference latency and energy per scene.}
\label{tab:runtime}
\adjustbox{max width=\textwidth}{
\begin{tabular}{m{.33\textwidth} m{.33\textwidth} m{.33\textwidth}}
\toprule
\textbf{Platform} & \textbf{Frame Rate (fps)} & \textbf{Energy (mJ)} \\
\midrule
RTX 4090 & 764.8 & 228.7 \\
Jetson AGX Xavier & 47.9 & 1218.5 \\
\textbf{ZCU102 FPGA} & \textbf{7} & \textbf{113.6} \\
\bottomrule
\end{tabular}
}
\end{table}
%
Table~\ref{tab:res_util} summarizes ZCU102 resource usage for the TinyIceNet accelerator.
%
\begin{table}
\centering
\caption{ZCU102 FPGA resource utilization.}
\label{tab:res_util}%
    \adjustbox{max width=\textwidth}{
    \begin{tabular}{m{.25\textwidth} m{.25\textwidth} m{.25\textwidth} m{.25\textwidth}}
    \toprule
    \textbf{Block RAM} & \textbf{DSP} & \textbf{Flip-Flops} & \textbf{Look-Up Tables} \\
    \midrule
    261 (14.31\%) & 121 (4.8\%) & 69,010 (12.59\%) & 156,575 (57.13\%)\\
    \bottomrule
\end{tabular}
}
\end{table}

\subsection{Comparison with Related Works}

To assess the effectiveness and efficiency of TinyIceNet, we compare it with existing SAR-only SOD segmentation models, specifically the MMSeaIce U-Net~\cite{chen2023mmseaice} and the DeepLabv3+ ResNet model~\cite{pires2023enhancing}. The comparison focuses on model size (number of parameters), computational complexity (MACs), SAR-only F1 score, and inference efficiency.

Table~\ref{tab:related_comparison} summarizes the comparison. TinyIceNet achieves competitive accuracy (F1=75.216\%) using only dual-polarized SAR inputs, closely matching the SAR-only ablation of MMSeaIce (F1=75.1\%) while requiring fewer parameters (146.6k vs 488.2~k) and lower computational cost (2.97~GMac vs 10.11~GMac) for $2\times512\times512$ input scenes. In contrast, DeepLabv3+ ResNet, although capable of SAR-based segmentation, is significantly larger (7M parameters) and computationally more expensive, with lower or variable multi-class F1 (60–80\%) and designed primarily for offline processing.

\begin{table}[ht]
\centering
\caption{Comparison of TinyIceNet with SAR-only SOD segmentation methods.}
\label{tab:related_comparison}
\adjustbox{max width=\textwidth}{
\begin{tabular}{m{.4\textwidth} m{.15\textwidth} m{.15\textwidth} m{.2\textwidth} m{.15\textwidth}}
\toprule
\textbf{Model} & \textbf{Parameters} & \textbf{MACs} & \textbf{SAR-only F1 (\%)} & \textbf{Frame Rate} \\
\midrule
MMSeaIce U-Net~\cite{chen2023mmseaice} & 488.2~k & 10.11~G & 75.1 & 0.5~fps \\
DeepLabv3+ ResNet (SAR-only)~\cite{pires2023enhancing} & 5.61M & 10.21~G & 60–80 & offline \\
\textbf{TinyIceNet (ours)} & \textbf{146.6~k} & \textbf{2.97~G} & \textbf{75.216} & \textbf{7~fps}\\
\bottomrule
\end{tabular}
}
\end{table}

\subsection{Qualitative Results}
\begin{figure}
    \centering
    \includegraphics[width=1\linewidth]{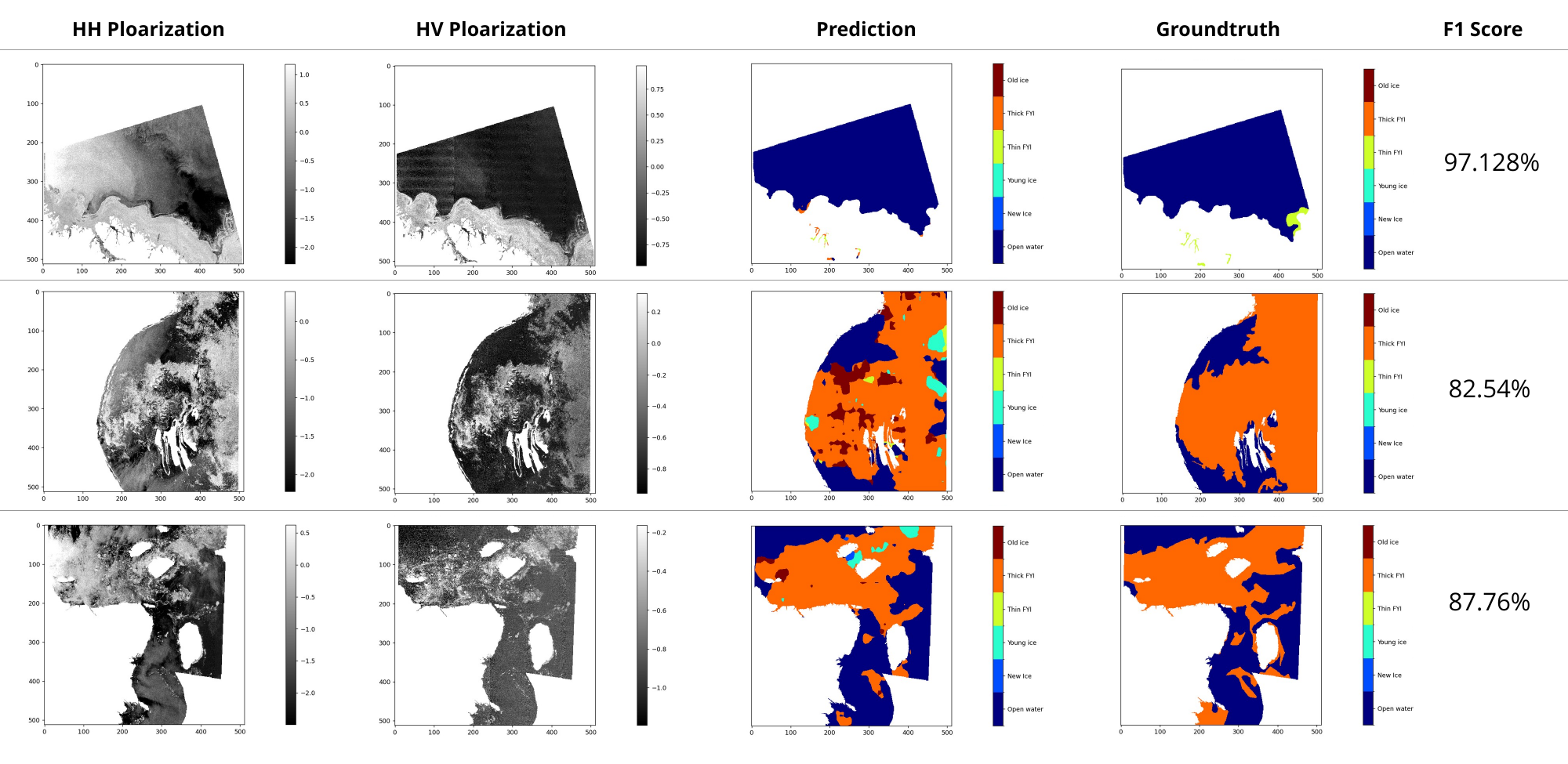}
    \caption{Qualitative comparison of TinyIceNet predictions. The individual F1 scores for the selected scenes are mentioned besides each inference row.}
    \label{fig:qualitative}
\end{figure}

Figure~\ref{fig:qualitative} presents qualitative examples of TinyIceNet predictions on three representative Sentinel-1 SAR scenes from the test set. Each row shows: (i) the HH polarization channel, (ii) the HV polarization channel, (iii) the predicted SOD map produced by the quantized TinyIceNet model, and (iv) the corresponding ground-truth annotation. For each example, the per-scene F1 score is reported to provide a quantitative reference for prediction quality.
The predicted maps preserve the overall morphology of ice floes and successfully separate open water, thin ice, and older ice types. Minor discrepancies are observed primarily near class boundaries, where the SAR signatures exhibit gradual transitions or low contrast. These boundary uncertainties are expected given the inherent ambiguity in SAR backscatter characteristics for adjacent SOD classes.

\section{Discussion and Future Works}
\label{section:conclusion_future_works}
The experimental results demonstrate that TinyIceNet is an effective and hardware-efficient solution for automated SOD mapping from Sentinel-1 SAR imagery.
By jointly considering network architecture, quantization strategy, and FPGA deployment constraints, the proposed approach achieves a favorable balance between accuracy, latency, and power consumption.
The integration of TinyIceNet on a Xilinx Zynq UltraScale+ MPSoC using the DeepEdgeSoC toolchain enabled an efficient translation from a high-level deep learning model to a fully functional hardware accelerator, delivering near--real-time inference with substantially lower energy consumption compared to GPU-based solutions.

Despite these promising results, several limitations of the current study should be acknowledged.
First, the evaluation is restricted to Sentinel-1 SAR data and to a single sea ice parameter (SOD), which limits the generality of the conclusions.
Second, although the AI4Arctic dataset spans multiple years and regions, variations in seasonal conditions, ice regimes, and geographical characteristics may affect model generalization beyond the tested scenarios.
From a hardware perspective, the observed degradation under aggressive post-training quantization highlights an inherent trade-off between numerical precision and accuracy, emphasizing the importance of task-specific quantization strategies for SAR data.

Future work will therefore extend this research along several complementary directions.
From a methodological perspective, the proposed co-design framework can be expanded to multi-task sea ice mapping, incorporating additional parameters such as sea ice concentration (SIC) or floe size (FLOE), as well as exploring multi-modal inputs that combine SAR with complementary data sources when hardware constraints permit.
From a hardware--algorithm co-design viewpoint, future efforts will investigate adaptive and mixed-precision arithmetic, as well as more advanced quantization-aware training strategies, to further reduce resource utilization while preserving robustness across diverse ice conditions.
Finally, deploying the system on different satellite or airborne platforms and validating its performance under real operational conditions will provide valuable insight into scalability, robustness, and practical usability, strengthening the path toward real-world adoption.
\section*{Acknowledgment}
This work has been partially funded by the European Union as part of the research project dAIEDGE (Grant number 101120726).
Furthermore, this work has been partially funded by the Federal Ministry of Education and Research of the Federal Republic of Germany as part of the research project FAIRe (Grant number 01IS23074).
\bibliographystyle{ieeetr}
\bibliography{Sections/99_references}


%
%
%







\end{document}